\documentclass[10pt,twocolumn,letterpaper]{article}

\usepackage{cvpr}
\usepackage{times}
\usepackage{epsfig}
\usepackage{graphicx}
\usepackage{amsmath}
\usepackage{amssymb}
\usepackage{algorithm}
\usepackage{algpseudocode}

\newcommand{\argmax}[1]{\underset{#1}{\operatorname{arg}\,\operatorname{max}}\;}

\usepackage[breaklinks=true,bookmarks=false]{hyperref}

\cvprfinalcopy % *** Uncomment this line for the final submission

\def\cvprPaperID{} % *** Enter the CVPR Paper ID here

\begin{document}

%%%%%%%%% TITLE
\title{Self-supervised visual feature learning with curriculum}
\author{Vishal Keshav\\
University of Massachusetts, Amherst\\
Massachusetts, 01002\\
{\tt\small vkeshav@cs.umass.edu}
\and
Fabien Delattre\\
University of Massachusetts, Amherst\\
Massachusetts, 01002\\
{\tt\small fdelattre@cs.umass.edu}
}

\maketitle
%\thispagestyle{empty}

%%%%%%%%% ABSTRACT
\begin{abstract}
Self-supervised learning techniques have shown their abilities to learn meaningful feature representation. This is made possible by training a model on pretext tasks that only requires to find correlations between inputs or parts of inputs. However, such pretext tasks need to be carefully hand selected to avoid low level signals that could make those pretext tasks trivial. Moreover, removing those shortcuts often leads to the loss of some semantically valuable information. We show that it directly impacts the speed of learning of the downstream task. In this paper we took inspiration from curriculum learning to progressively remove low level signals and show that it significantly increase the speed of convergence of the downstream task. Code is made publicly available at \url{https://github.com/vishal-keshav/self-supervised-curriculum}.

\end{abstract}

%%%%%%%%% BODY TEXT

\section{Introduction}

Computer vision tasks including object classification~\cite{classification}, detection~\cite{detection} and segmentation~\cite{segmentation} are dominated by supervised learning approaches, mostly because of the presence of large scale annotated datasets like Image-net~\cite{ILSVRC15} and Open Images~\cite{OpenImages2}  Dataset. However, most of the unlabelled visual data like images and videos are available in copious amount and labeling them is a tedious and complex process, fetching supervised methods impractical. To make sense of diverse yet highly structured visual data, a new class of algorithm called self-supervised learning~\cite{self} methods tends to exploit the spatial context present in the images and videos by treating them as a source of free and abundant amount supervisory signals. For example, self-supervised techniques like Jigsaw puzzle solver~\cite{jigsaw}, colorization~\cite{colorization}, distortion ~\cite{dosovitskiy2015discriminative} and rotation~\cite{rotation} learn Spatio-temporal correlation present in the image by solving a predefined task (called pretext task). These tasks assist in extracting a meaningful representation of the data without explicitly needing any labels (where labels are algorithmically generated). Rich representation captured is then used to aid the learning process of a data-scarce downstream task such as classification, detection or segmentation.\\

\begin{figure*}[htbp]
\begin{center}
%\fbox{\rule{0pt}{2in} \rule{.9\linewidth}{0pt}}
\includegraphics[width=0.9\linewidth]{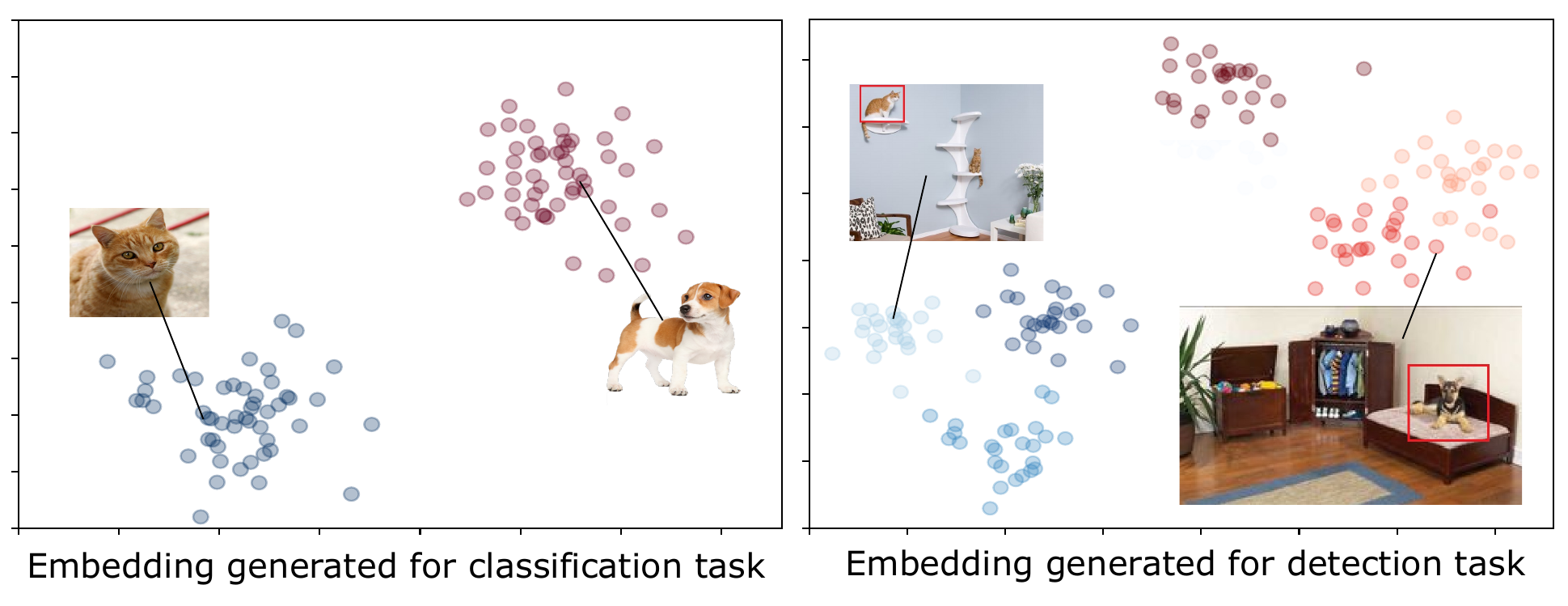}
\end{center}
   \caption{In this figure, we demonstrate an ideal learned embedding produced with a self-supervised learning technique and plot them in a low dimensional manifold. On the left, the learned embedding can be a good candidate for classification tasks because it was able to disentangle the clusters for each classes. On the right, the learned embedding is suitable for detection tasks as it encodes the relative position of an object in an image along with class labels.}
\label{fig:cluster}
\end{figure*}

These pretext tasks, however, do not guarantee that they learn a robust representation as is demanded by several downstream tasks. We say a feature is robust if the captured embedding manifests intra-class variation while remaining invariant to inter-class dissimilarity. For a classification task, a robust representation for images will lead to highly clustered low-dimensional embeddings where the individual clusters represent the individual classes needed to be classified. For a localization or detection task, a robust representation will result to a cluster of clusters, where the ultimate cluster constitutes of embedding representing the relative location of the object in an image. Fig.~\ref{fig:cluster} illustrates this concept visually.\\

% [Vishal] Uptil here, I am satisfied. Refined the caption for first figure.

%[Vishal] Do you suggest an alternative to the term heuristics? Or shall we go with it?

To learn a robust representation of images concerning a downstream task, several transformations in pretext task training procedures have been applied. For example, in jigsaw puzzle solver and patch-based reordering~\cite{patch}, it was observed that the network was able to find a shortcut by learning the relative order of the texture-based image patches through chromatic aberration\cite{waller2010phase} present in the image. To avoid that, the authors applied the shifting of color channels and then jittering the color channels independently. Such a data transformation made the task tougher to learn, but at the same time avoided the shortcuts that a network can learn to solve the pretext task. We discuss the concept of task difficulty and shortcuts learned by a pretext task learner in the next section.\\

%[Vishal] The below paragraph can be improved for clarity, but not neccessary.
In this paper, we explore the critical factors that construct a self-supervised pretext task avoid shortcuts while making the learning process less ambiguous. With our empirical evaluations and the understanding of the shortcut avoiding factors essential for a given self-supervised pretext task, we propose a general learning framework for self-supervised learning which is based on curriculum learning, enabling us to learn a robust set of features for classification downstream task.\\

The paper is organized as follows: Section 2 provides a detailed review of several self-supervised tasks for visual feature learning. We develop some terminologies we use in the rest of the paper. Section 3 formally describes the problem statement. Section 4 discusses our approach. Section 5 demonstrates the experimental results with our approach. Finally, we conclude the research with our findings in section 6.

\section{Background/Related Work}
In deep learning literature, convolutional neural networks~\cite{lecun2015deep}~\cite{khan2019survey} have been acclaimed as the best method to learn hierarchical features from image and video data, where low-level features represent lines and other primitive shapes while the high-level features correspond to high-level semantic features that combine the low-level features. When the data is scarce, these methods fail to accomplish this hierarchical feature learning as they tend to overfit on the dataset. Self-supervised learning methods combined with deep architectures helps overcome this issue by learning low-level to medium level features without annotated data and providing the learned parameters as a good starting point to the original task to be learned. The learned features can be transferred, finetuned or can be used to generate a better and easy to learn representation for the scarce dataset.\\

Below we describe a generic self-supervised learning procedure to learn a task $T$ with limited annotated dataset $X = \{(x_{1},y_{1}), (x_{2},y_{2}), ..., (x_{n},y_{n})\}$:
\begin{enumerate}
    \item Design a pretext task $T_{p}$.
    \item Collect a large collection of un-annotated but related data points, call it $X_{p} = \{ \hat{x}_{1},\hat{x}_{2} ..., \hat{x}_{m}\}$ and generate the labels $\hat{y}_{p} = \{ \hat{y}_{1}, \hat{y}_{2}, ..., \hat{y}_{m}\}$ based of designed task $T_{p}$ where $m>>n$.
    \item Train the task $T_{p}$ on dataset $(X_{p},\hat{y}_{p})$ and retain the features learned from the training procedure in terms of $\phi$.
    \item Train the original downstream task $T$ on the featurized dataset $\{ (\phi(x_{1}), y_{1}), (\phi(x_{2}), y_{2}), ..., (\phi(x_{n}), y_{n})\}$.
\end{enumerate}
In the next two subsections, we provide a detailed description of two pretext tasks that learns the latent structure present in the image data using spatial contextual cues.

%[Vishal] Uptill this point, I think we are good.

\subsection{Jigsaw Puzzle Solver}
The main idea behind the Jigsaw puzzle solver is to teach a network that an object is made up of parts and the semantics associated with these building blocks. The authors of this paper achieve this goal by cropping an image, dividing the crop into several patches, shuffling the patches using a predefined permutation, and then feeding the shuffled patches into a network. The network is then trained to recognize the permutation used to shuffle the patches. The network learns to disentangle the factors of variation in order to correctly predict the permutation of the shuffled images. These factors are likely be the inherent spatial structure present in the objects of the image. Since the labels (the permutations) can be algorithmically generated, the dataset required for this pretext task is clearly un-annotated. One more thing to observe here is that they use a single network where the weights are shared across each head (that consumes one of the shuffled image patches). Once the network produces the embedding for each of the patches, those embeddings are then concatenated and processed with a fully connected layer to predict the shuffling order.\\

\begin{figure*}[htbp]
\begin{center}
%\fbox{\rule{0pt}{2in} \rule{.9\linewidth}{0pt}}
\includegraphics[width=0.9\linewidth]{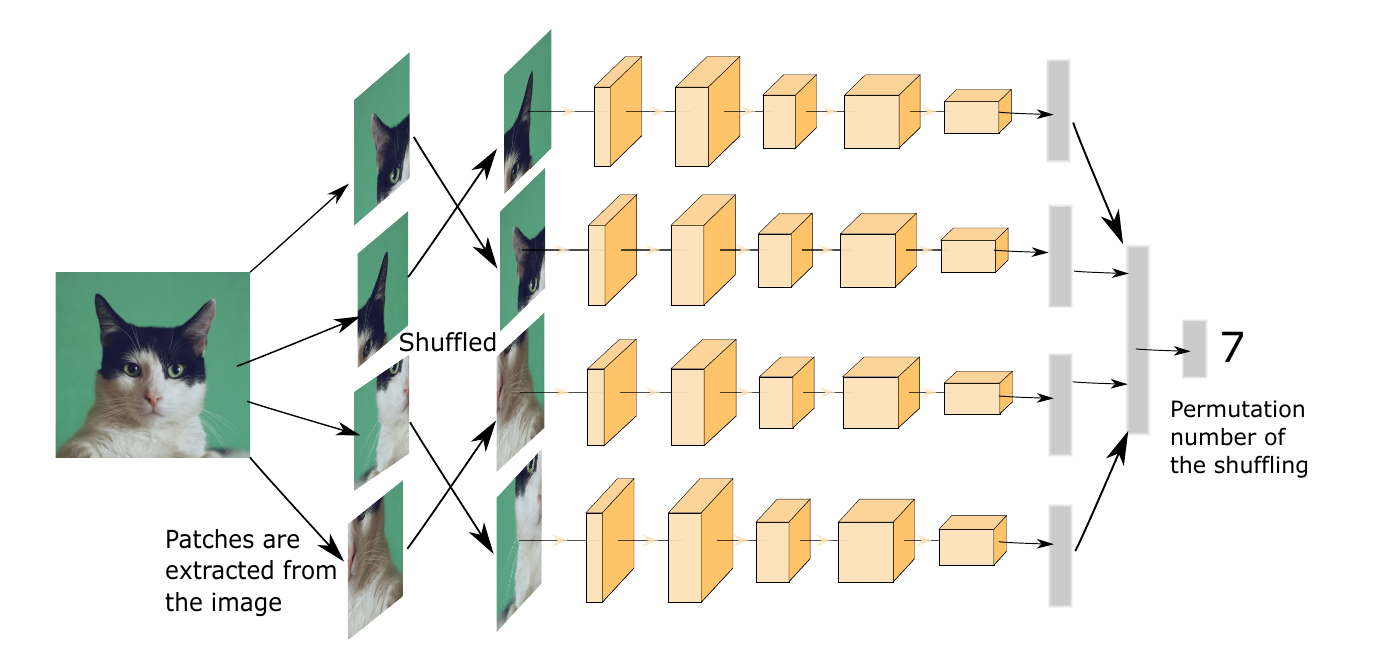}
\end{center}
   \caption{This figure shows the method employed by jigsaw puzzle solver. An input image is first cropped into four quadrants. Then a permutation of 4 is randomly selected and patches are shuffled according to the permutation. The patches are fed to the network where the weights (as shown by the yellow tensors) are shared. The output embedding are then concatenated to generate a permutation prediction.}
\label{fig:jigsaw}
\end{figure*}

In the original experiments, the authors chose to work with 3X3 patches, with which 9! possible permutations can be generated to shuffle the image patches. To keep the search space low, they generate a subset of these 9! permutations ensuring that the hamming distance between the selected permutations is high. In our testbed, we chose to experiment with 4 image patches, which can be permuted with at-most 4! = 24 permutations. In Figure \ref{fig:jigsaw}, we show an  overview of this method.

\subsection{Image-patch reordering}
Similar to the jigsaw puzzle, the patch-based relative position predictor task divides a cropped image into 3X3 image patches. But unlike jigsaw puzzle solver, a random pair of these 3X3 patches are fed to the network. The network is then trained to predict the correct ordering in relation to the central patch. The intuition is that while the network learns to predict the relative ordering, it extracts essential semantic information that constitutes those images. In their work, the authors use a Siamese like network with shared weights. Just like a jigsaw puzzle solver, the embedding from the network is concatenated and processed with a fully connected layer. The output from this layer reflects one of the 8 possible positions of one of the image patch in relation to the central patch. Since this task also relies on algorithmically cropping the image and feeding a random pair, the supervisory signals can be automatically generated.\\

\begin{figure*}[htbp]
\begin{center}
%\fbox{\rule{0pt}{2in} \rule{.9\linewidth}{0pt}}
\includegraphics[width=0.9\linewidth]{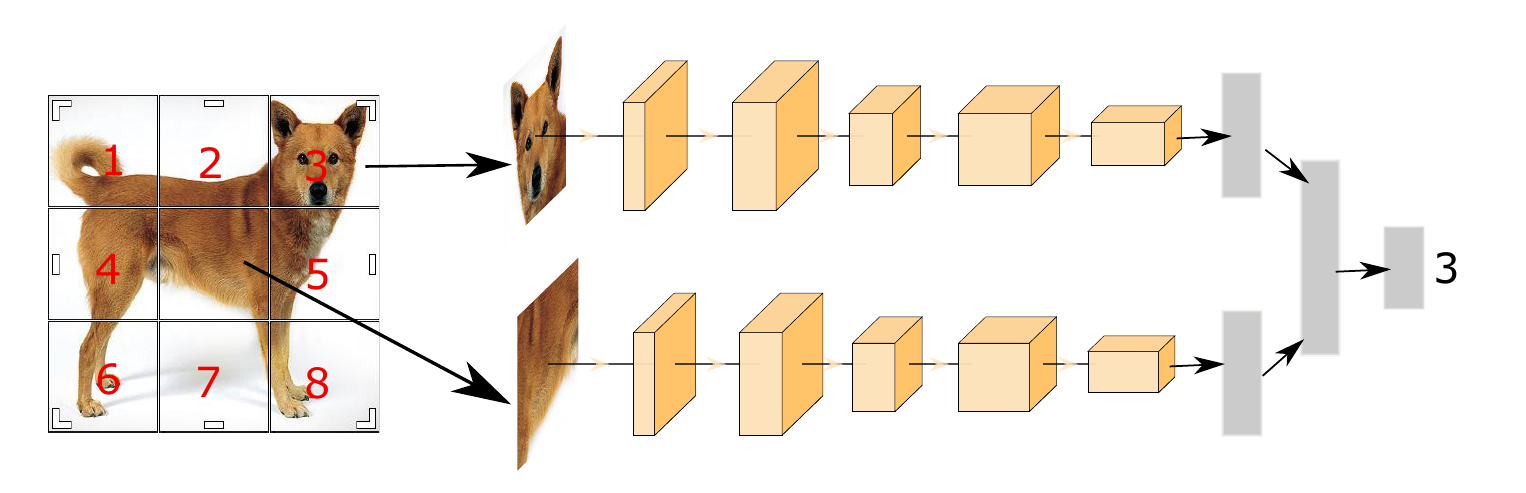}
\end{center}
   \caption{The patch based re-ordering task requires the network to learn relative ordering of image patches. To do this, an image is first cropped into 3X3 equal parts, then a random cropped patch is selected in relation to the central patch. The patches are fed in a network that share the parameters. The output embedding is then concatenated and prediction is made, which reflects the position of the first patch in relation to the second.}
\label{fig:patch}
\end{figure*}

Although we conduct multiple experiments with this task, we validate our primary approach only with the jigsaw puzzle solver. The reason being that this task performed not as good as the jigsaw puzzle solver. We believe that the representation captured through this pretext task is not as effective as the jigsaw puzzle solver for classification downstream tasks. In Figure~\ref{fig:patch}, we showcase the overview of this method.\\

%[Vishal] "at hand", phrase means that "the one we are currently concerned with".

\subsection{Shortcuts and difficulty in self-supervised learning}
\begin{figure}[!htb]
\begin{center}
   \includegraphics[width=0.9\linewidth]{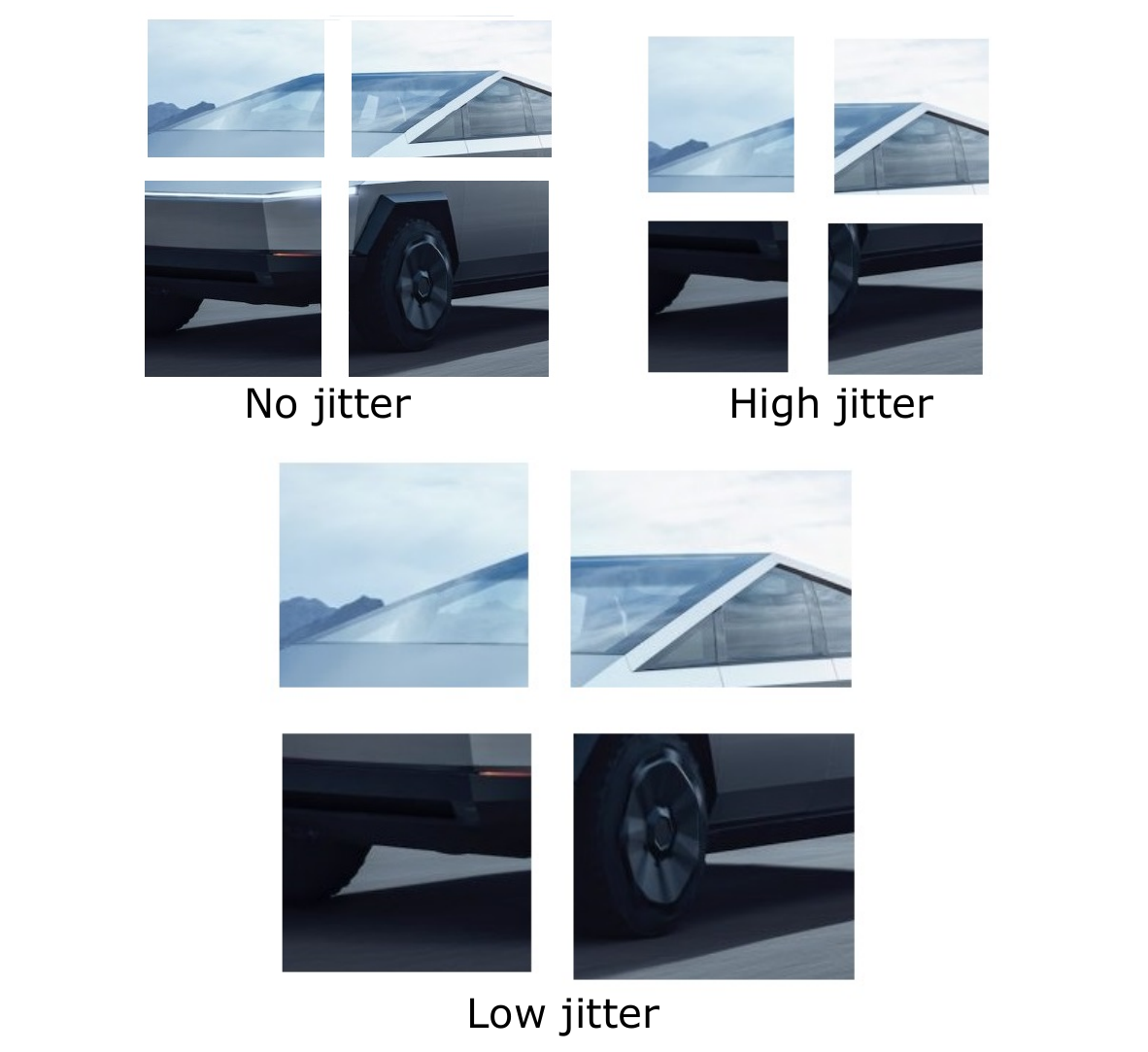}
\end{center}
   \caption{In this figure, we show the effect of applying varying degree of jitters on the cropped patches. Low jitter will constitutes high semantic value, but network would learn taking the shortcuts based on neighboring patch patterns. High jitter will avoid those shortcomings but the semantic value will be lost.}
\label{fig:jitter}
\end{figure}

One of the drawbacks of above mentioned self-supervised methods, as observed by the respective authors was that the network tried to take shortcuts by picking up low-level signals from the images to learn the pretext task at hand. As a result, such learning did not fetch any good representation as required for the downstream task training. The low-level features learned by the network such as detecting color aberration and patch wise neighboring pixel correlation have no value over semantically rich features. To avoid these issues, the authors proposed data transformation techniques such as applying jitter to the image patch, randomly greyscaling the patches to remove color abbreviation and patch wise image normalization. These heuristics helped overcoming the shortcut learning but at the same time, the application of these transformations lead to the loss of some of the semantically valuable information present in the image. A visual depiction of the effect of applying jitter is shown in Figure~\ref{fig:jitter}. While, it is true that introducing such transformation makes the task avoid taking shortcuts, but those transformations also  degrade the quality of the learned features because of the loss in the semantically relevant information.
%[vISHAL] ->    Rephrased this red line: { \color{red}While, it is true that a difficult task captures rich features, but at the same time the quality of these learned features will degrade because of the loss in the semantically relevant information}.\\

In this paper, we tend to resolve this issue by introducing a generic training method for self-supervised learning. In the next section, we formally describe the problem statement.

%[Vishal] : Uptill here, we are good.

%[Vishal]: I think we can remove the curriculum learning section because we will describe these before our approach. What do you suggest? Its too early to reveal on the key ingridient presented in the paper. We should keep the interest up atleast till approach section,
%\subsection{Curriculum learning}

%{\color{blue}Curriculum learning~\cite{bengio2009curriculum} is directly inspired by how humans and animals learn. Early research~\cite{elman1993learning} have shown that humans and animals starts by learning simple concepts and then use them to learn more complicated ones. This same idea has been successfully applied to deep learning ~\cite{graves2017automated}~\cite{murali2018cassl}. The model is first train on easy examples and then train on more complex ones. It has been shown that this could speed up convergence and improve generalization. We propose to use curriculum learning on the pretext task to remove progressively the potential shortcuts and thus keeping all the structural information at the beginning of the training.}

\section{Problem Formulation}
The main idea presented in the paper is to incrementally increase the difficulty of the pretext task to reduce the initial information loss. We formally present the problem statement as follow:\\
Let $X^{1}_{p}, X^{2}_{p}, ..., X^{k}_{p}$ represents a sequence of data-distribution in increasing difficulty order to train a pretext task $T_{p}$, where $X^{1}_{p}$ is the easiest set of data points to train the task $T_{p}$ whereas $X^{k}_{p}$ represent the most difficult data points (with possibly lesser semantic concepts). Let us define a function $f$ that measures the difficulty of training distribution $X^{i}_{p}$. We say $X^{i}_{p}$ is easier than $X^{j}_{p}$ if $f(X^{i}_{p}) < f(X^{j}_{p})$. Furthermore, let us define $\phi^{i}$ to be the feature learned from training the task $T_{p}$ over data distribution $X^{i}_{p}$. If $(X,y)$ is the original dataset on which downstream task $T$ is needed to perform well, our objective is to maximize the test accuracy of task $T$. In other words, we want to train pretext task $T_{p}$ on data distribution $X^{i}_{p}$ where $i$ is given by following equation:
\begin{equation}
    i =  \argmax{i}\frac{1}{n}\sum(T(\phi^{i}(X)) == y)
\end{equation}

This objective, however, cannot be easily optimized. For lower $i$, the examples are easier and the network will find shortcuts. For larger $i$, most of the semantic information will be lost. In both of these cases, we end up with a sub-optimal feature representation. Hence, we want to combine the data distributions with varying difficulty that can fetch us a robust representation $\phi$. Next, we present our approach to achieve the desired result.

% [Vishal] -> Uptill this point, we are good.

\section{Approach}
The solution to the above-mentioned problem is presented in two folds. First, we find the critical factors required to make a self-supervised training tasks learn robust features, and then we use the factor to device a principled learning strategy to train the task $T_{p}$ such that the test accuracy of the task $T$ over the learned features $(\phi(X),y)$ can be maximized.\\

\subsection{The critical factor to improve jigsaw puzzle solver}
We conduct a comprehensive review to determine the factors that improve the features learned by jigsaw puzzle solver. As indicated by M. Noroozi et al.\cite{jigsaw}, the network learns shortcuts to make the self-supervised task learn better. We heed towards the factors which helped resolve the shortcoming of the learning process in the jigsaw puzzle solver and find the most influential ones among all the transformations applied by the authors:\\
\begin{itemize}
    \item \textbf{Patch-based normalization:} Once the image has been cropped into four patches, each patch is individually normalized. The normalization makes the patches statistically independent. For our purpose, we apply a normalization with a mean 0 and a standard deviation of 1.0. We point out that this transformation did not affect the effectiveness of feature learning.
    \item \textbf{Random greyscaling:} The images are randomly greyscaled with a probability of 0.3. Since the network may learn the color aberration present in the image, this transformation effectively removes such a shift in color channels. In our evaluations, this transformation improved the accuracy of the downstream task but by only a marginal amount.
    \item \textbf{Random jitter:} The image patches are randomly cropped to jitter the image patches. In our evaluation of the factors reducing the shortcuts, we set the random cropping to 95\% of the total patch portion. Above or below 95\% of random cropping did not produce good results.
\end{itemize}

Our evaluations of these data transformations are detailed in the experiment section. We found that random jitter transformation heuristic boosted the downstream classification task accuracy more than any other discussed transformations.\\

%[Vishal] -> {\color{red} improvise} is not a correct term, I wrote "devise" instead.
%[Vishal] -> Shall we remove section 2.4 bacause I took the content you wrote there and put it in apporach section. The reason being, I want the paper to feel interesting till the reader reads the approach section. Otherwise, it becomes to obvious what we are trying to do.
% [Vishal] -> I have colored the section in blue and removed {\color{red} this is said in 2.4}
\subsection{Feature learning with curriculum}
We discuss our approach to devise a better self-supervised feature learning strategy commonly applied to supervised training called curriculum learning~\cite{bengio2009curriculum}. Curriculum learning is inspired by the idea of how humans learn~\cite{elman1993learning}. To learn a complex concept, a curriculum involves learning first the easy task and then keep on increasing the difficulty level of the concept. Similar to this, machine learning systems which when exposed to easier examples first and then tuned towards difficult ones, performs better. Curriculum learning have been successfully applied to deep learning ~\cite{graves2017automated}~\cite{murali2018cassl} to learn hierarchically complex features. The intuition behind this learning strategy is that it changes the landscape of the optimization procedure making it easy for the learner to reach a better local optimum. \\

Inspired by the idea, we propose to incorporate the curriculum learning to train the self-supervised pretext task, which is the jigsaw puzzle. We first define some terminologies, then provide the complete algorithm for the proposed learning strategy. Let us call $g$ be the data transformer that takes dataset and one other argument(representing the difficulty) used to transform the data. For example, in jigsaw puzzle solver, we choose jitter levels as possible set of second argument where $g$ applies the corresponding jitter level to the input dataset. The algorithm is as follow:\\

\begin{algorithm}
	\caption{Curriculum for self-supervision}
	\begin{algorithmic}[1]
	    \State \textbf{Input:} dataset \textbf{$X_{p}$}, task \textbf{$T_{p}$}, jitter levels \textbf{$J$}
	    \State \textbf{Output:} trained task $T_{p}$
		\For {$i=1,2,\ldots |J|$}
            \State Generate	$X^{i}_{p}$ = $g(X_{p}, J[i])$
		\EndFor
		\State Sort $X^{i}_{p}$ such that $f(X^{i}_{p}) < f(X^{j}_{p})$ for $\forall$ $i<j$.
		\For {$i=1,2,\ldots |J|$}
            \State Train task $T_{p}$ on $X^{i}_{p}$
		\EndFor
		\State Return the trained task $T_{p}$
	\end{algorithmic}
\end{algorithm}

Intuitively, this learning procedure helps the pretext task to first acquire the semantics of the input. Along with it, the task also adapts to take shortcuts. As we increase the jitter level, the learner is provided with lesser cues for the shortcut, but get to utilize the semantics learned from previous training epochs. This enables the learner to evolve the features which are more robust for a downstream task. In the next section, we demonstrate the effectiveness of the trained task $T_{p}$ using the above algorithm.

%[Vishal] -> Uptill this point, I think everythin looks good.

\section{Experiment}
\subsection{Dataset}
\begin{figure*}[htbp]
\begin{center}
%\fbox{\rule{0pt}{2in} \rule{.9\linewidth}{0pt}}
\includegraphics[width=0.9\linewidth]{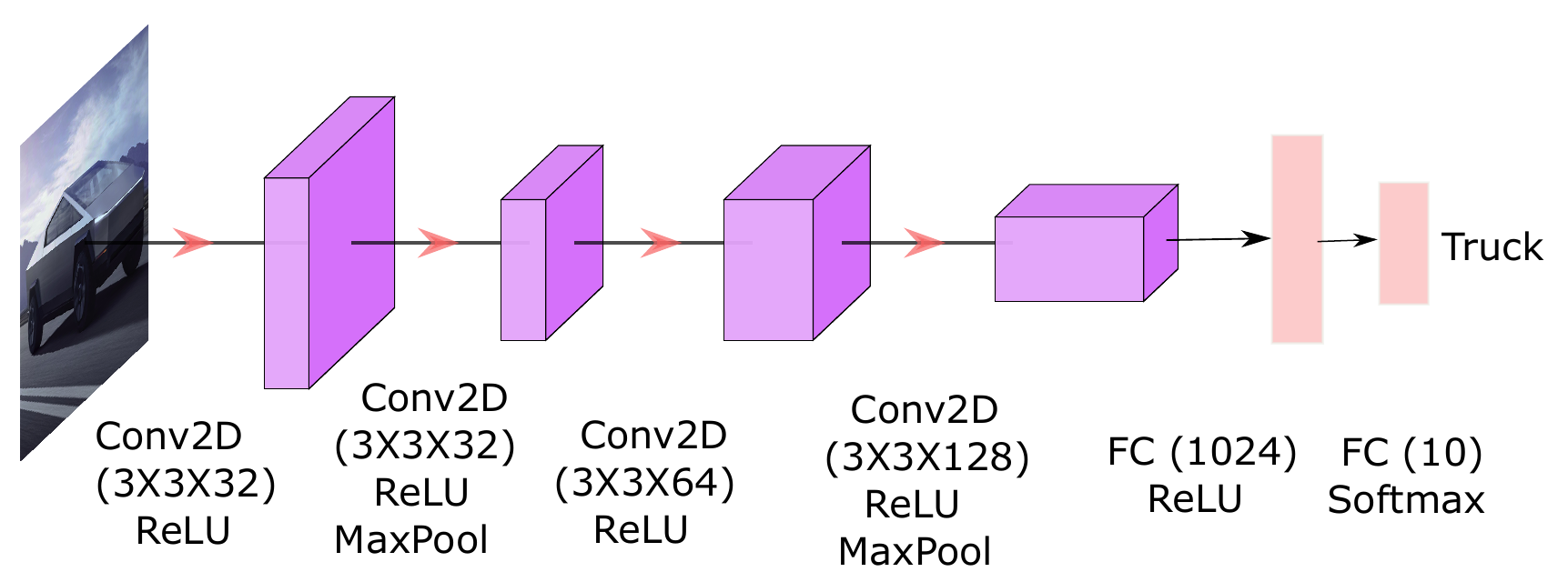}
\end{center}
   \caption{The network for the classification task is shown. Both the pretext task and the downstream task use the same architecture. For classification task, we remove the fully connected layer from the network used for pretext task, and add fully connected layer that outputs the matching dimension for the number of classes to be classified.}
\label{fig:arch}
\end{figure*}

We conduct all our experiments on the STL10 image recognition dataset~\cite{coates2011analysis}. The characteristics of the dataset are provided in table \ref{table:dataset}. Each image is of size 96X96 and there are 10 different classes in the dataset. The pretext task is trained on the unlabelled dataset and tested on the test set (where we remove the labels). The downstream task is trained on the labeled train dataset and tested on the labeled test dataset.  The downstream task is trained with the same weights as that of a trained pretext task network. The network architecture used for the downstream classification task is shown in Figure~\ref{fig:arch}.

\begin{table}[ht]
\centering
\caption{Characteristics of the STL10 image dataset for unsupervised visual learning.}
\begin{tabular}[t]{lc}
\hline
Split Category &Number of datapoints\\
\hline
Unlabeled images &  100000  \\
Labeled train images & 500  \\
Labeled test images & 800 \\
\hline
\label{table:dataset}
\end{tabular}
\end{table}%

%[Vishal] -> Uptill here everything looks good.

\subsection{Experiments conducted to find the influential factors for jigsaw puzzle task}
\begin{figure}[!htb]
\begin{center}
   \includegraphics[width=0.9\linewidth]{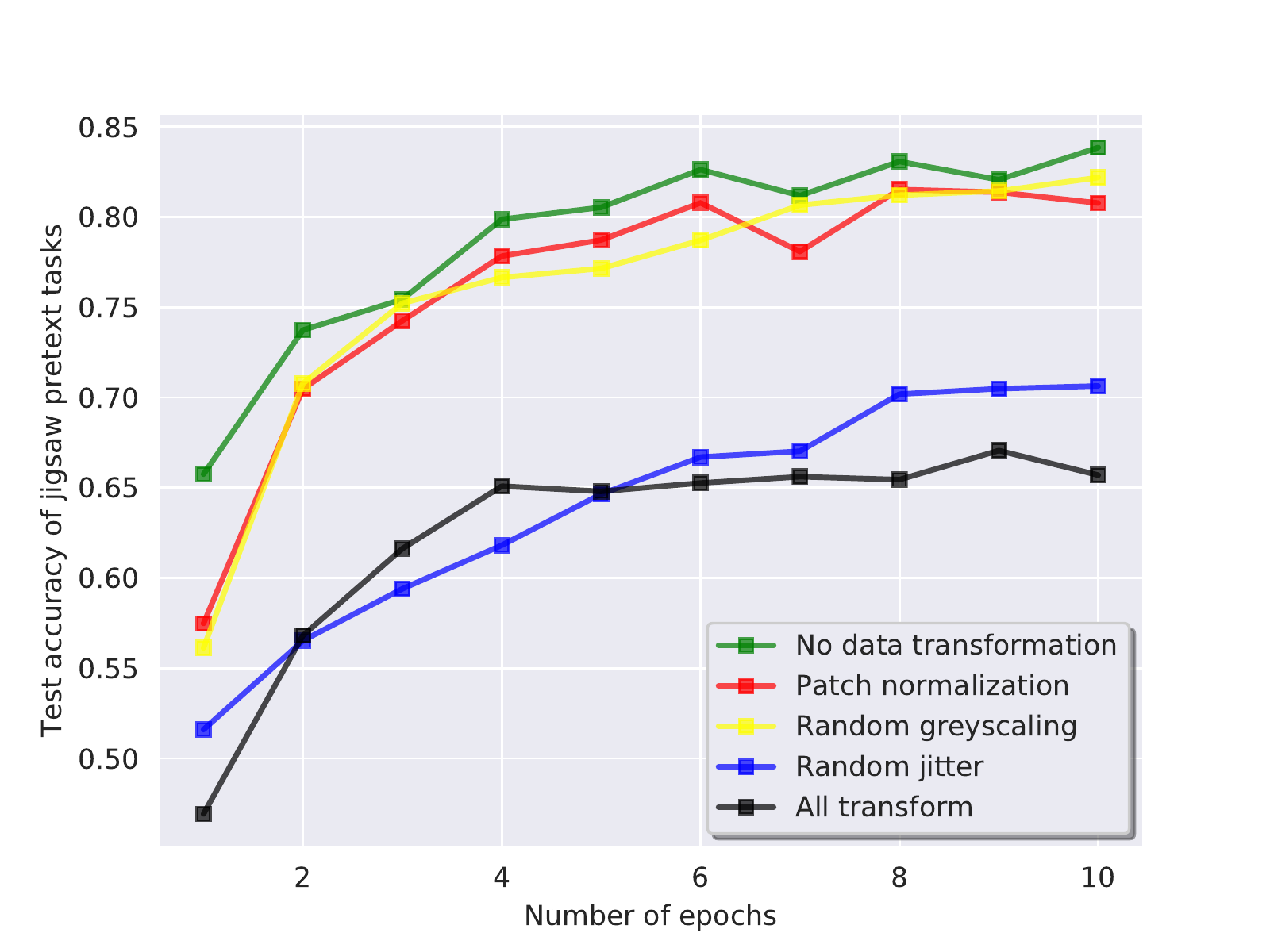}
\end{center}
   \caption{Jigsaw solver pretext task test accuracy. The accuracies are shown with respect to the number of training epochs.}
\label{fig:influential_plot}
\end{figure}

We train the jigsaw puzzle solver task and image classification task with STL10 dataset and empirically determine which transformation worked best for the pretext task learning and which one resulted in a good representation for the downstream classification task. The data transformation includes applying independent path normalization, random jitter and random greyscaling. We greyscale the images with a probability of 0.3. Jittering is applied by randomly cropping 95\% of the patch regions. The plot in Fig.\ref{fig:influential_plot} shows the test accuracies of the pretext task network after applying the above-mentioned transformation.\\

\begin{figure}[!htb]
\begin{center}
   \includegraphics[width=0.9\linewidth]{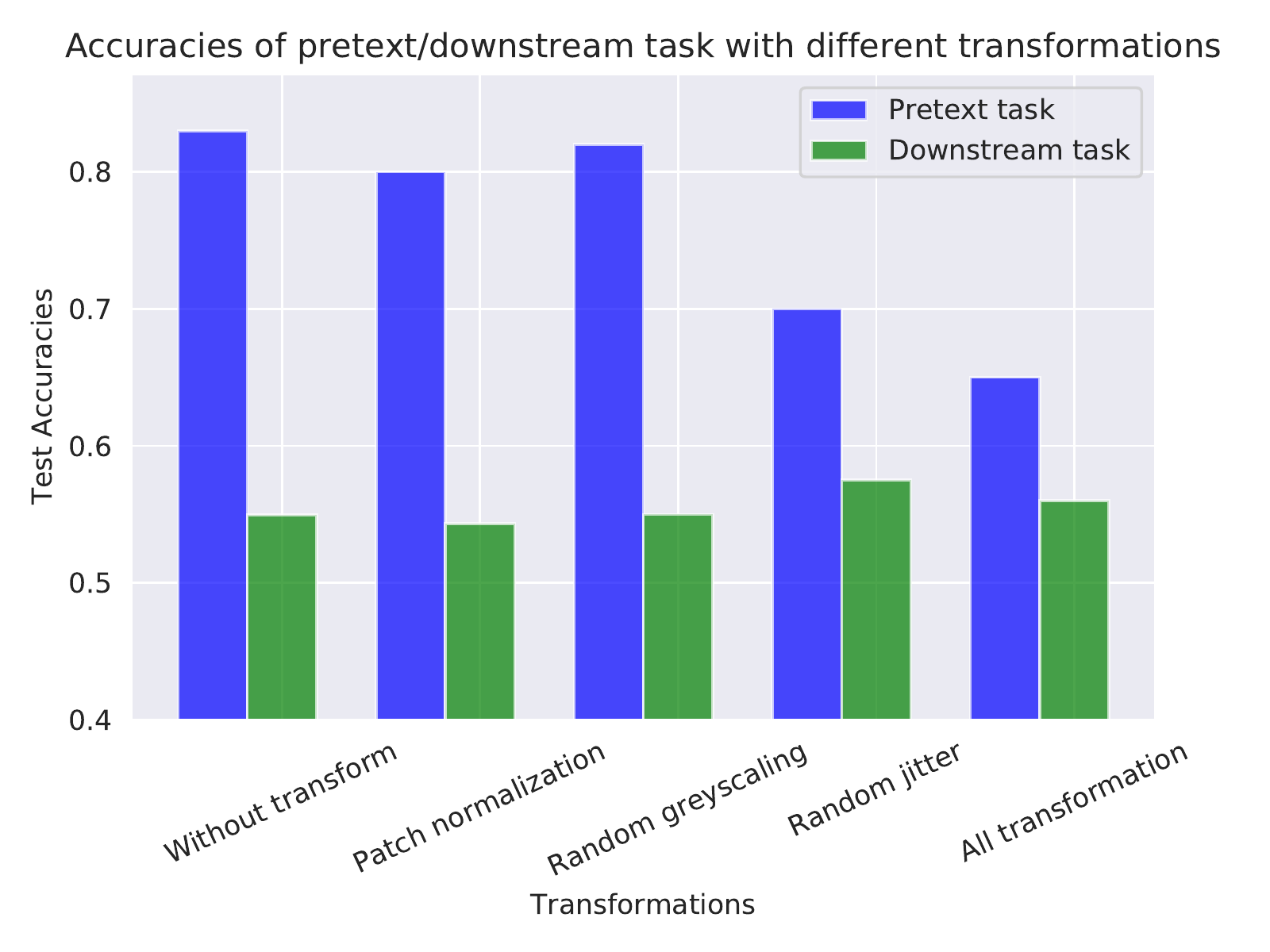}
\end{center}
   \caption{Jigsaw solver pretext task test accuracy and the accuracy on the classification task using the learned weights from the jigsaw training. We compare the accuracy to showcase the quality of learned features with pretext task training. Better the classification accuracy, robust will be the learned representation. The results are averaged over 10 runs.}
\label{fig:influential_bar}
\end{figure}

In Figure \ref{fig:influential_bar}, we show the test accuracies of pretext task and downstream classification tasks with different transformations. It can be observed that after the jitter transformations, the network performed better than any other transformation method when it came to learning a good representation for the classification task. Another startling thing to note here is that when we apply all the mentioned transformations to the data and then train the pretext task, both pretext task accuracy and downstream task accuracy goes down as compared to applying only random jitter. This strongly suggests that there is no correlation between the difficulty attained by a pretext task and the quality of features acquired for the downstream classification task.

\subsection{Experimental results with the proposed learning strategy}
In this section, we provide the accuracies gained with our proposed training approach. The experimental setup is as follow:\\
\begin{enumerate}
    \item First, we choose the jitter as the only influencing factor to learn a good representation.
    \item For curriculum learning, we vary the jitter from no cropping to randomly cropping 80\% of the image patch, with a step size of 3\% to 5\%.
    \item For base comparisons, we fix the jitter constants to 100\%, 95\%, 90\%, and 80\% and training the pretext and corresponding downstream task.
\end{enumerate}

\begin{figure}[!htb]
\begin{center}
   \includegraphics[width=0.9\linewidth]{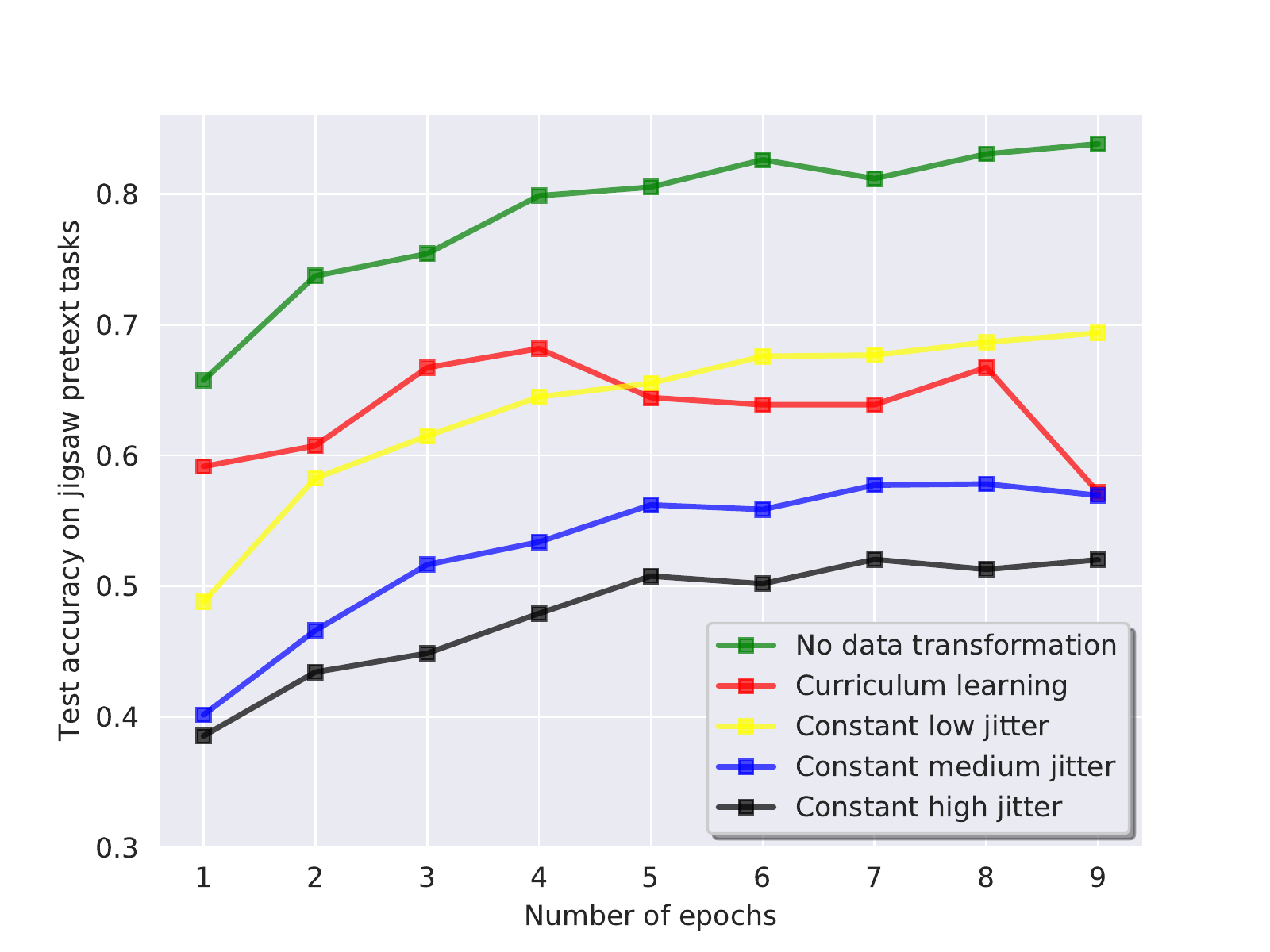}
\end{center}
   \caption{The plot shows the pretext task test accuracies achieved with curriculum learning and with constant jitter levels. As can be seen, the curriculum learning strategy progressively makes the task difficult to learn.}
\label{fig:curriculum_plot}
\end{figure}

\begin{figure}[!htb]
\begin{center}
   \includegraphics[width=0.9\linewidth]{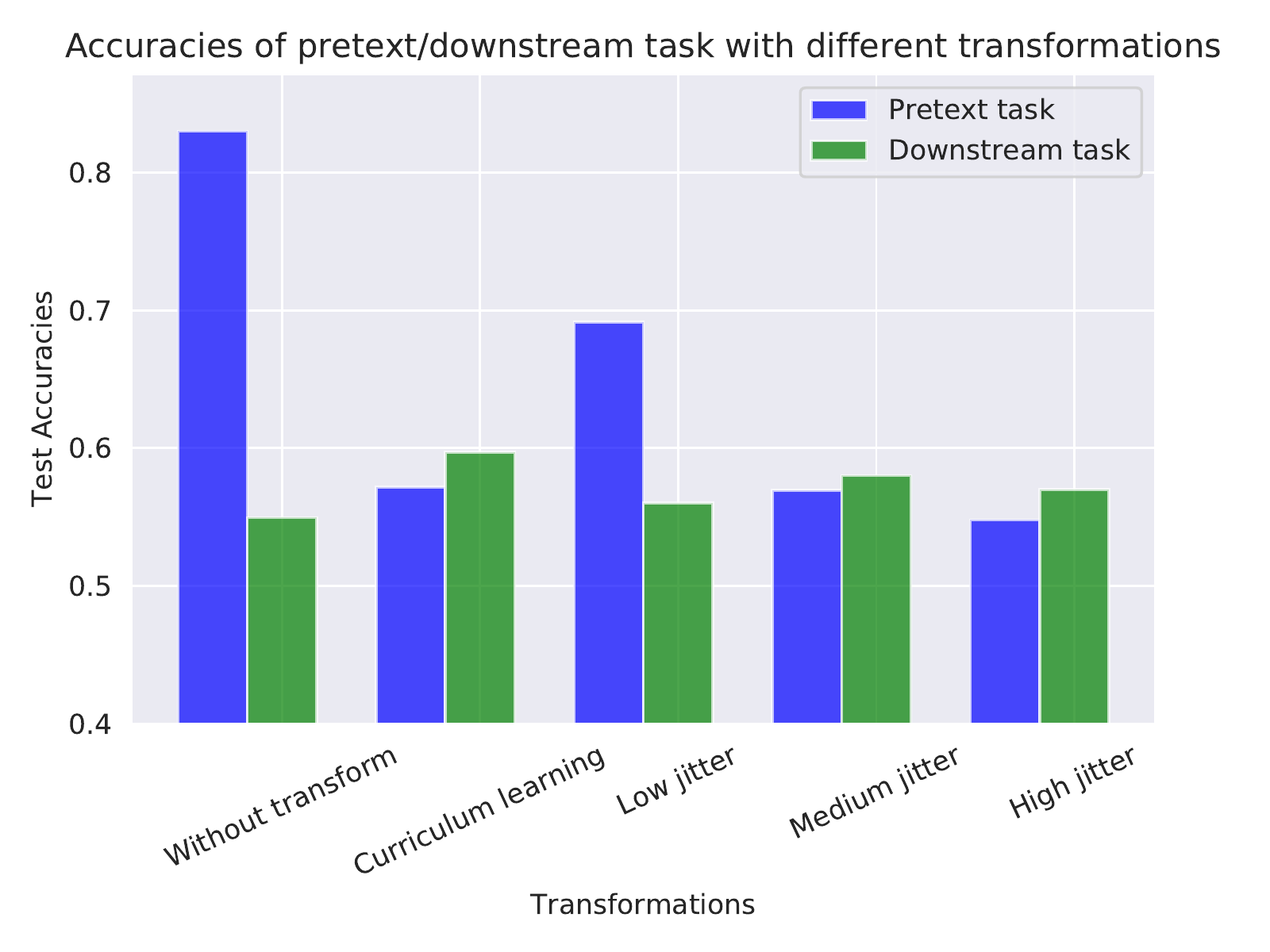}
\end{center}
   \caption{The bar plot compares the achieved test accuracy with curriculum learning and with different jitters values. The curriculum learning approach tends to learn a rich set of features as demonstrated by higher downstream task accuracy. The results are averaged over 10 runs.}
\label{fig:curriculum_bar}
\end{figure}

In Fig. \ref{fig:curriculum_plot}, we show the pretext task accuracies with varying levels of jitter along with the curriculum learning technique. The plot suggests that as we progress with the training, curriculum learning makes the task difficult to learn. As per our claim, the bar plot in Fig. \ref{fig:curriculum_bar} reflects that the features learned in such a fashion produces a better representation. This is demonstrated by comparing the accuracy level of the downstream task with and without curriculum learning.

\subsection{Implementation details}
We train the models on both pretext tasks and downstream classification tasks using the architecture shown in Figure \ref{fig:arch}. All the implementations were done in Pytorch. We use a learning rate of 0.001 along with Adam optimizer for parameter updates. We use 12 different permutations for jigsaw puzzle tasks and 10 classes for the classification task. Networks were trained on a single Nvidia GTX 1080 Ti GPU. Code is made publicly available at \url{https://github.com/vishal-keshav/self-supervised-curriculum}.\\

%[Vishal] -> This section needs paraphrasing a little bit
\subsection{On direct feature evaluation metrics}
\begin{figure}[!htb]
\begin{center}
   \includegraphics[width=0.8\linewidth]{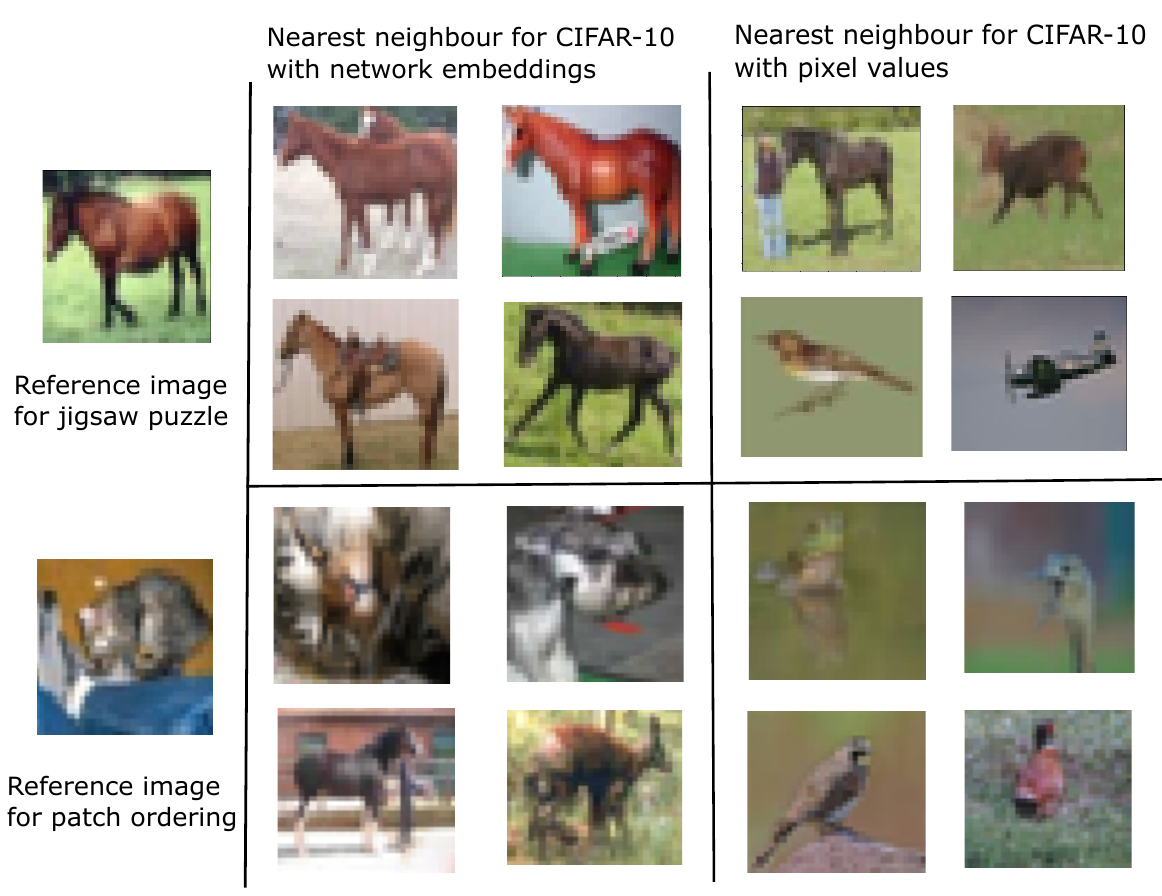}
\end{center}
   \caption{Nearest neighbours based on the distance in embedding and pixel space for self-supervised tasks, namely jigsaw solver and patch based ordering on a sample of images.}
\label{fig:nearest_neighbors}
\end{figure}

A typical method employed to evaluate how good are the features obtained through self-supervised training is to find the nearest neighbors and see if they belong to the same class. In Fig. \ref{fig:nearest_neighbors}, we show the nearest neighbors we obtained on CIFAR-10 dataset~\cite{krizhevsky2009learning}. In the absence of a qualifying downstream task, it becomes difficult to directly measure the effectiveness of how good were the learned features with self-supervision.\\

%[Vishal]: Till here, we are good.

%[Vishal]: This conclusion looks bad. Can you help in refining this? Just like you did with the abstract.
\section{Conclusion}
In this paper, we study the factors responsible for making a self-supervised task to learn a efficient representation for downstream task. Towards this end, we conduct a comprehensive survey on several pretext tasks designed to capture spatial contextual information from the images and use the learned features to improve classification task accuracy. We conduct exhaustive experimentation to study the trade-off between the shortcuts learned by the network in order to solve the pretext task and the semantics required to learn a good feature representation for downstream task. We use this fact to employ the curriculum learning technique to induce a better feature representation for the final downstream task. The results on the classification task with our proposed changes strongly suggest that the learned parameters capture a rich set of representation required for the classification problem. With the obtained results, we also conclude that there is no direct relationship between pretext task accuracy and the corresponding downstream task accuracy.\\
As a future research direction, we tend to generalize the training strategy we introduced for jigsaw puzzle solver for other pretext tasks.\\

%[Vishal]: Uptill here, everything looks fine.

% We study the trade-off between the shortcuts learned by the network in order to solve the pretext task and the semantics required to learn a good feature representation for downstream task.

% We do exhaustive experiments on the data transformation

% We study the trade-off between keeping all the semantic informations of the images and avoiding shortcuts.

{\small
\bibliographystyle{ieee}
\bibliography{egbib}
}

\end{document}